\documentclass[runningheads]{llncs}
\usepackage[T1]{fontenc}
\usepackage{graphicx}
\usepackage{amsmath} 
\usepackage{amssymb}
\usepackage[misc]{ifsym}

\begin{document}
\title{BCRNet: Enhancing Landmark \\Detection in Laparoscopic Liver Surgery \\via Bezier Curve Refinement}
\titlerunning{BCRNet for Liver Landmark Detection}

\author{Qian Li\inst{1} \and
Feng Liu\inst{2} \and
Shuojue Yang\inst{1} \and
Daiyun Shen\inst{1}\and
Yueming Jin\inst{1}\textsuperscript{(\Letter)}
}
\authorrunning{Q. Li et al.}
% First names are abbreviated in the running head.
% If there are more than two authors, 'et al.' is used.
%
\institute{
National University of Singapore, Singapore \\
\email{ymjin@nus.edu.sg}
\and 
Harbin Institute of Technology, Harbin, China
}

\maketitle              % typeset the header of the contribution
\begin{abstract}

Laparoscopic liver surgery, while minimally invasive, poses significant challenges in accurately identifying critical anatomical structures. Augmented reality (AR) systems, integrating MRI/CT with laparoscopic images based on 2D-3D registration, offer a promising solution for enhancing surgical navigation. A vital aspect of the registration progress is the precise detection of curvilinear anatomical landmarks in laparoscopic images. In this paper, we propose \textbf{BCRNet} (\textbf{B}ezier \textbf{C}urve \textbf{R}efinement \textbf{Net}), a novel framework that significantly enhances landmark detection in laparoscopic liver surgery primarily via the Bezier curve refinement strategy. The framework starts with a Multi-modal Feature Extraction (MFE) module designed to robustly capture semantic features. Then we propose Adaptive Curve Proposal Initialization (ACPI) to generate pixel-aligned Bezier curves and confidence scores for reliable initial proposals. Additionally, we design the Hierarchical Curve Refinement (HCR) mechanism to enhance these proposals iteratively through a multi-stage process, capturing fine-grained contextual details from multi-scale pixel-level features for precise Bezier curve adjustment. Extensive evaluations on the L3D and P2ILF datasets demonstrate that BCRNet outperforms state-of-the-art methods, achieving significant performance improvements. Code will be available.

% Authors must provide keywords and are not allowed to remove this Keyword section.

\end{abstract}

\section{Introduction}
Minimally invasive surgery (MIS) has gained widespread clinical adoption due to its advantages over traditional open surgery, including reduced trauma and faster recovery \cite{fuchs1998augmented}. However, laparoscopic images under MIS present inherent challenges such as a narrow field of view, frequent occlusion and difficulty in spatial orientation, bringing difficulty for surgeons in recognizing anatomical structures or tumours. Augmented reality (AR) could mitigate these issues by overlaying preoperative 3D MRI/CT data onto intraoperative surgical view \cite{ramalhinho2023value}. %, enhance anatomical visualization . %The core of AR implementation lies in the 
Accurate detection of landmarks in 3D MRI/CT data and 2D laparoscopic images, such as ridge, falciform ligament, and silhouette for liver surgery,
can benefit precise 2D-3D registration, which serves as a crucial prerequisite in AR to enhance intraoperative visualization.

3D images could be precisely marked preoperatively, however, it is highly desired to develop automated and reliable algorithms to online detect the landmarks in 2D intraoperative laparoscopic images. Nevertheless, accurately identifying these landmarks in laparoscopic images remains challenging, due to complex surgical environments, interference from surgical actions, tissue deformation, bleeding, and instrument occlusion. % making it difficult for landmarks to be clearly presented and accurately identified. 

Most existing approaches treat landmark detection as a segmentation problem. Previous study \cite{labrunie2022automatic} has utilized a U-Net model for dilated liver landmark segmentation. However, limitations in semantic understanding have resulted in numerous predictions falling outside acceptable error ranges. Subsequent work by \cite{labrunie2023automatic} have improved landmark perception through co-attention mechanisms, yet challenges remain. More recently, D$^2$GPLand\cite{pei2024depth} introduced the Segment Anything Model (SAM) with depth information and furthered segmentation accuracy. Despite these advancements, those pixel-level classification methods still lead to issues like curve discontinuities and isolated segmentation pixels, particularly in complex cases. Additionally, these methods often require post-processing to recover landmark coordinates and do not explicitly determine the order of points along landmark curves, complicating subsequent 2D-3D registration.

Bezier curve technique, an advanced curve detection method, has gained increasing research interest and shown promising performance in applications within the natural domain, such as text recognition \cite{liu2021abcnet,ye2023deepsolo} and lane detection \cite{feng2022rethinking,blayney2024bezier}. These methods are considered more precise than segmentation-based approaches, as segmentation focuses on local features, while Bezier curves offer a holistic representation of curvilinear landmarks \cite{feng2022rethinking}. However, surgical images, with greater interference and complexity than natural images, present significant challenges. Existing methods struggle with complex surgical landmark detection, yet their achievements inspire exploration in surgical scenarios.

In this paper, we propose BCRNet, an innovative framework that utilizes Bezier curves to align complex semantic information in surgical laparoscopic images, enabling accurate liver landmark detection. To address the challenges of complex surgical scenarios, BCRNet introduces several distinct innovations.
Our model begins with a \textit{Multi-modal Feature Extraction} (MFE) module designed to integrate both visual and crucial geometric information. Augmented by advanced vision foundation models, it significantly enhances the accuracy of curve detection for liver landmarks by extracting richer high-level semantic features. Liver landmark curves are then encoded as efficient Bezier curves, ensuring continuity and alignment with landmark definitions. The framework integrates an \textit{Adaptive Curve Proposal Initialization} (ACPI) module and a \textit{Hierarchical Curve Refinement} (HCR) module for curve initialization and refinement. The ACPI module generates a pixel-aligned Bezier control points map and a confidence score map, enabling systematic selection of optimal curve proposals. Furthermore, we introduce a Proposal Induction Loss which accelerates training convergence by guiding the model more effectively in the systematic selection of optimal curve proposals. Subsequently, to specifically prevent our model from converging to local minima due to poor initializations which is a significant risk in this challenging task, the HCR module employs a coarse-to-fine strategy. It iteratively enhances these proposals by progressively leveraging global-to-local contextual information. Moreover, by partially using features at each stage, HCR allows for more refinement iterations within the same computational budget, thereby improving accuracy. Extensive experiments on two public datasets, L3D and P2ILF, demonstrate the superiority of our model over other state-of-the-art methods.

\section{Methods}
\begin{figure}[t]
    \centering
    \includegraphics[width=0.93\textwidth]{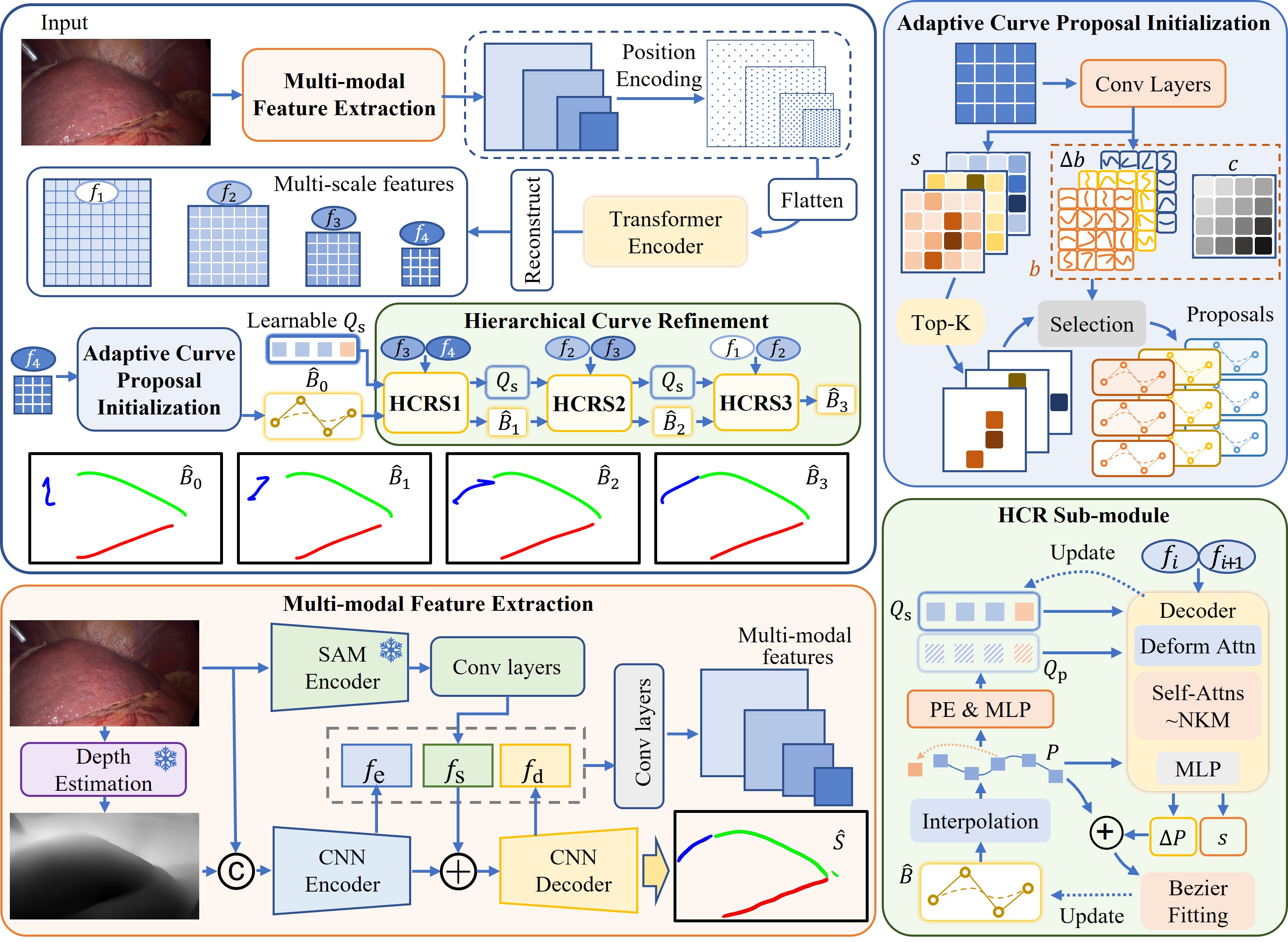} 
    \caption{Overview of the proposed BCRNet for liver landmark detection.  The framework starts with MFE to capture semantic features. ACPI then generates pixel-aligned Bezier curves and confidence scores, enabling systematic initialization of curve proposals. These proposals are subsequently refined by HCR, leveraging progressively detailed representations for precise curve adjustment.}
    \label{fig:method}
\end{figure}

In this study, we address the challenge of detecting curvilinear liver landmarks by employing Bezier curves as an efficient and accurate representation method. Our proposed network architecture, illustrated in Figure \ref{fig:method}, firstly comprises a MFE module (Sec. \ref{sec:2.2}) and a transformer encoder to extract a series of multi-scale features \( f_i, i=1,2,3,4 \) from the input images. 
For each landmark categroy, the ACPI module (Sec. \ref{sec:2.3}) generates Bezier curve proposals from the feature map \( f_4 \). Each curve is characterized by six Bezier control points and a corresponding confidence score.
Subsequent refinement is achieved through the HCR module (Sec. \ref{sec:2.4}), which progressively optimizes the curve proposals by leveraging more detailed multi-scale features.

\subsection{Multi-modal Feature Extraction}
\label{sec:2.2}
To exploit the geometric properties in liver landmarks, input laparoscopic images are first augmented with depth information using AdelaiDepth \cite{yin2022towards}. The obtained RGB-D images are then processed by a CNN encoder to extract spatial features, denoted as $f_\text{e}$. Additionally, leveraging the robust performance of SAM \cite{kirillov2023segment}, a frozen SAM encoder extracts supplementary features, $f_\text{s}$, from the RGB image for anatomical structure identification. The features from both encoders are fused to form a comprehensive multi-modal representation. To enhance the explicitness of these features for downstream tasks, we introduce a CNN decoder, which is trained specifically for landmark segmentation. This decoder refines the features and guides the model to learn more interpretable representations, producing decoder features $f_\text{d}$. Features from the encoders ($f_\text{e}$ and $f_\text{s}$) and decoder ($f_\text{d}$) are integrated across scales, yielding the extracted multi-modal features represented with multi-scales.

\subsection{Adaptive Curve Proposal Initialization}
\label{sec:2.3}
To enhance the representation of curvilinear liver landmarks and improve detection performance, this study introduces a novel liver landmark presentation scheme. Each landmark is mathematically modeled as a fifth-order Bezier curve, parameterized by six 2D control points. This approach effectively accommodates a variety of landmark shapes while utilizing a minimal number of control points. To address the inherent uncertainty in the number of curves for each landmark category, our framework adopts a multi-proposal prediction strategy for robust detection. It enhances the system's adaptability by enabling the generation of variable curves for each landmark and providing the flexibility to predict several segments when landmarks are partially occluded.

As illustrated in Figure \ref{fig:method}, the ACPI module utilizes the $f_4$ feature, which encapsulates rich global information, to generate landmark curve proposals. Inspired by previous work \cite{ye2023deepsolo}, we predict a set of residual offsets for each pixel. By integrating these predicted residual offsets $\Delta b$ with the normalized coordinate map $c$ of pixel indices, we obtain a Bezier curve control points map $b \in \mathbb{R}^{H\times W\times 12}$. Each pixel in this map represents the coordinates of six control points, precisely and efficiently defining the shape and position of the predicted curve. Specifically, for the $i$-th pixel with normalized coordinates $c_i=(c_{ix},c_{iy})$, an offset vector $\Delta b_i=[\Delta b_{ix}^{1}, \Delta b_{iy}^{1},\Delta b_{ix}^{2},\Delta b_{iy}^{2},...,\Delta b_{iy}^{6}]$ is predicted. The Bezier control points $b_i$ can then be calculated, with the coordinates of the $j$-th point expressed as follows:

\begin{equation}
    b_i^j=(\sigma(\Delta b_{ix}^j+\sigma^{-1}(c_{ix})), \sigma(\Delta b_{iy}^j+\sigma^{-1}(c_{iy})))
\end{equation}
where $\sigma$ denotes the sigmoid function. Additionally, a corresponding proposal score map $s$ is predicted to indicate the confidence of the pixel-aligned Bezier curves. For each landmark category, the top-$K$ scored curves $\hat{B}_0=\{b_k|k \in \text{topk}(s)\}$ are selected as the initial proposals.

\subsection{Hierarchical Curve Refinement}
\label{sec:2.4}
The curves predicted by the ACPI module often lack sufficient detail to accurately delineate target landmarks, necessitating refinement through more informative guidance. To address this, we propose a HCR module based on the deformable cross-attention mechanism \cite{zhudeformable}. It consists of three sub-modules (HCRS1, HCRS2, and HCRS3 in Figure \ref{fig:method}) to adjust the curves. The refinement process follows a hierarchical structure, processing semantic features from $\{f_3, f_4\}$, $\{f_2, f_3\}$, and $\{f_1, f_2\}$ in sequence, which allows for gradual adjustments of the Bezier control points and their associated proposal scores, significantly enhancing curve accuracy and precision.

Within each module, inspired by previous work \cite{ye2023deepsolo}, we uniformly sample $N-1$ curve points $P_s$ on each Bezier curve. We also introduce a global point $P^*$ from the middle of the curve, to capture global information. This results in normalized reference point coordinates $P=[P_s, P^*] \in \mathbb{R}^{M \times K \times N \times 2}$. Point positional queries $Q_p$ are generated through positional encoding and MLP layers: $Q_p = \text{MLP}(\text{PE}(P)) \in \mathbb{R}^{M \times K \times N \times 2}$. Furthermore, we introduce a learnable semantic query $Q_s$ to capture content-specific information. The composite queries $Q$ are then formed $Q = Q_p + Q_s$ for transformer decoder layers.

Within the decoder, queries undergo deformable cross-attention\cite{zhudeformable} to aggregate information from multi-scale features, using $P$ as reference points. A self-attention process is sequentially structured to further enhance information interaction, including (1) intra-curve attention across dimension $N$ to model relationships between points on the same curve; (2) inter-curve attention across $K$ captures relationships between proposals; and (3) inter-category attention across $M$ explores relationships between different landmark categories. 
After these operations, an MLP module predicts offsets $\Delta P_s$ for each sampled point and computes corresponding proposal scores $s$ by averaging the confidence scores of the $N$ reference points.

Finally, a new Bezier curve $\hat{B}$ is fitted from the updated points $P_s$ to regularize their shape and distribution.

\subsection{Optimization}
\label{sec:2.5}

Following previous work \cite{zhang2022text}, we employed a Bipartite matching strategy during network training to align predicted curves with Ground Truth (GT) for loss computation. A similar matching cost is utilized, in which we modify the curve distance term to simultaneously consider the distances on Bezier control points and interpolated curve points. For the predicted confidence scores, we used the same focal loss \(\mathcal{L}_\text{cs}\). When calculating the curve distance loss \(\mathcal{L}_\text{crv}\) , we account for both control points and interpolated curve points to ensure comprehensive geometric alignment.  Additionally, several auxiliary loss functions are designed to facilitate network convergence and enhance training stability.

\noindent\textbf{Semantic Segmentation Loss.} During the training of the MFE module, the Deep Supervision strategy is applied. The semantic segmentation results $\hat{S}_l$ ($l = 1, 2, 3, 4$) from the CNN decoder are utilized to compute the semantic segmentation loss $\mathcal{L}_s = \sum_{l=1}^4 \text{Dice}(\hat{S}_l, S)$.

\noindent\textbf{Proposal Induction Loss.} 
In early training stage, the network may produce inaccurate proposals and confidence scores, leading to inefficient GT-prediction matching via the injective function. Therefore, we introduce a proposal induction loss with BCE function: $\mathcal{L}_\text{ind} = \text{BCE}(\hat{s}_{\text{init}}, s^*)$, where $\hat{s}_{\text{init}} \in \mathbb{R}^{bs \times M \times H \times W}$ is the predicted proposal confidence score, and $s^* \in \{0, 1\}^{bs \times M \times H \times W}$ is the designed induction map, with 1s at GT landmark midpoints and 0s elsewhere.

\noindent\textbf{Overall Loss.} 
Through the proposal initialization and hierarchical refinement process, we obtain four sets of predictions from these stages, denoted as $\{\hat{B}_h, \hat{s}_h|h=0,1,2,3\}$, and a multi-level supervision strategy is utilized during training. Since $\mathcal{L}_s$ and $\mathcal{L}_\text{ind}$ aim to accelerate network convergence in the early stages, a decay coefficient $\lambda_{\text{d}}$ is introduced: $\lambda_{\text{d}} = 1-\sigma((epoch-10)/2)$. Thus, the overall loss function is defined as:
\[
\mathcal{L} = \lambda_{\text{d}} (\lambda_s \mathcal{L}_s + \lambda_\text{ind} \mathcal{L}_\text{ind}) + (1 - \lambda_{\text{d}}) \sum_{h=0}^3 \left(\lambda_\text{cs}\mathcal{L}_\text{cs}(\hat{s}_h) + \lambda_\text{crv}\mathcal{L}_\text{crv}(\hat{B}_h)\right).
\]

\section{Experiments and Results}
\subsection{Experimental settings}

\noindent\textbf{Datasets and Evaluation.} 
We evaluated BCRNet on two publicly available datasets: L3D \cite{pei2024depth} and P2ILF \cite{ali2025objective}. L3D contains 1,152 keyframe images from surgical videos of 39 patients, annotated with three types of semantic landmarks as polylines. This dataset is pre-divided into three subsets: a training set of 921 images, a validation set of 122 images, and a test set of 109 images.
P2ILF is designed for 2D/3D landmark detection and 2D-3D image registration, comprising 167 images, which we split into 71, 27, and 69 images as training, validation and test set, respectively. Landmarks are dilated 30 pixels for evaluation as the operations in D2GPLand\cite{pei2024depth}.
Due to its limited size, P2ILF is insufficient for training a robust model, so we evaluated models pre-trained on L3D and fine-tuned on P2ILF in subsequent experiments. Other methods are evaluated with the same protocol in P2ILF for fair comparison. Following \cite{pei2024depth}, we used Intersection over Union (IoU), Frequency-Weighted IoU (FWIoU), Dice Score Coefficient (DSC), and Average Symmetric Surface Distance (ASSD) as evaluation metrics. 

\noindent\textbf{Implementation Details.}
We use SAM-ViT-B, and a U-Net built upon ResNet50 in MFE. For the deformation attention module, we use 8 heads and 4 sampling points. In the ACPI module, 256 Bezier curves are predicted per landmark category, with \( K = 10 \) selected as proposals. The number of reference points \( N \) is set to 26. During evaluation, a proposal score threshold of 0.3 is applied to determine the final output. The loss weights \( \lambda_{\text{s}} \), \( \lambda_{\text{ind}} \), \( \lambda_{\text{cs}} \), and \( \lambda_{\text{crv}} \) are configured as 10.0, 1.0, 1.0, and 1.0, respectively. Training is conducted 60 epochs on the L3D dataset and 20 epochs for fine-tuning on P2ILF, with a batch size of 4. The Adam optimizer is employed with an initial learning rate of \( 1 \times 10^{-5} \) and a weight decay of \( 1 \times 10^{-4} \). All experiments are implemented in PyTorch and executed on a single NVIDIA RTX A6000 GPU.

\begin{table}[h]
    \centering
    \begin{minipage}{0.45\linewidth}
        \centering
        \resizebox{\textwidth}{!}{
        \begin{tabular}{p{2.8cm}p{1cm}p{1cm}p{1cm}}
        \hline
        Model         & DSC$\uparrow$  & IoU$\uparrow$   & ASSD$\downarrow$  \\ \hline
        Deepsolo++\cite{ye2023deepsolo}     & 62.43 & 48.21 & 65.81  \\
        ABCNet v2\cite{liu2021abcnet} & 53.40 & 38.13 & 78.49   \\ \hline
        Unet$^*$\cite{ronneberger2015u}          & 51.39 & 36.35 & 84.94   \\
        COSNet$^*$\cite{labrunie2023automatic}        & 56.24 & 40.98 & 69.22   \\
        ResUNet$^*$\cite{xiao2018weighted}       & 55.47 & 40.68 & 70.66   \\
        Unet++$^*$\cite{zhou2019unet}        & 57.09 & 41.92 & 74.31   \\
        HRNet$^*$\cite{wang2020deep}         & 58.36 & 43.50  & 70.02   \\
        DeepLabv3+$^*$\cite{chen2018encoder}    & 59.74 & 44.92 & 60.86   \\
        TransUNet$^*$\cite{chen2021transunet}     & 56.81 & 41.44 & 76.16   \\
        SwinUNet$^*$\cite{cao2022swin}      & 57.35 & 42.09 & 72.80    \\ \hline
        SurgicalSAM\cite{yue2024surgicalsam}  & 62.56 & 47.86 & 68.47   \\
        SAM-Adapter$^*$\cite{wu2023medical}   & 57.57 & 42.88 & 74.31   \\
        SAMed$^*$\cite{zhang2023customized}         & 62.03 & 47.17 & 61.55   \\
        SAM-LST$^*$\cite{chai2024ladder}       & 60.51 & 45.03 & 68.87   \\
        AutoSAM$^*$\cite{hu2023efficiently}       & 59.12 & 44.21 & 62.49   \\
        D$^2$GPLand\cite{pei2024depth}      & 64.04 & 49.54 & 60.80    \\
        BCRNet & \textbf{69.57} & \textbf{54.16} & \textbf{43.55}  \\ \hline
        \end{tabular}
        }
        \caption{Quantitative comparison with SOTAs on the L3D dataset. $^*$ Note: Results are derived from \cite{pei2024depth}.}
        \label{tab:comparison_l3d}
    \end{minipage}
    \hspace{0.01\linewidth}
    \begin{minipage}{0.45\linewidth}
        \centering
        \resizebox{\textwidth}{!}{
            \begin{tabular}{p{2.8cm}p{1cm}p{1cm}p{1cm}}
            \hline
            Model        & DSC$\uparrow$  & IoU$\uparrow$   & ASSD$\downarrow$   \\ \hline
            Deepsolo++\cite{ye2023deepsolo}     & 48.34 & 34.82 & 138.46 \\
            ABCNet v2\cite{liu2021abcnet} & 42.32 & 31.41 & 184.65 \\
            Surgical SAM\cite{yue2024surgicalsam}  & 46.53 & 33.61 & 164.56 \\
            D$^2$GPLand\cite{pei2024depth}      & 48.73 & 35.52 & 132.10  \\
            BCRNet & \textbf{56.96} & \textbf{40.33} & \textbf{89.69}  \\ \hline
            \end{tabular}
        }
        \caption{Quantitative comparison with SOTAs on the P2ILF dataset. Results were obtained from models  fine-tuned on the P2ILF dataset. }
        \label{tab:comparison_p2ilf}
        \vspace{1em} 

        \resizebox{\textwidth}{!}{
            \begin{tabular}{ccccccc}

            \hline
            Method & MFE              & HCR                       & ACPI                      & $\mathcal{L}_\text{ind}$ & DSC$\uparrow$  & FWIoU$\uparrow$   \\ \hline
            M.1     &                     &                    &                   &                                                           & 38.77 & 22.46 \\
            M.2     & \checkmark &                   &                    &                                                           & 40.13 & 22.81 \\
            M.3     & \checkmark & \checkmark &                   &                                                           & 53.67 & 34.87 \\
            M.4     & \checkmark & \checkmark & \checkmark &                                                           & 66.32 & 46.92 \\
            M.5     &                     & \checkmark & \checkmark & \checkmark                                 & 65.73 & 47.06 \\
            Ours   & \checkmark & \checkmark & \checkmark & \checkmark                                 & \textbf{69.57} & \textbf{49.86} \\ \hline

            \end{tabular}
        }
        \caption{Quantitative analysis of the key components on the L3D dataset. }
        \label{tab:ablation}
    \end{minipage}
    \vspace{-3mm}
\end{table}

\subsection{Comparison with State-of-the-Art Methods}

We compared the proposed BCRNet with 16 SOTA methods on the L3D dataset, including Bezier-curve-based detection methods (Deepsolo++ \cite{ye2023deepsolo}, ABCNet v2\cite{liu2021abcnet}), non-SAM-based segmentation models (UNet\cite{ronneberger2015u}, COSNet\cite{labrunie2023automatic} , ResUNet\cite{xiao2018weighted} , DeepLabV3+\cite{chen2018encoder} , UNet++\cite{zhou2019unet} , HRNet\cite{wang2020deep}  , TranUNet\cite{chen2021transunet} , SwinUNet\cite{cao2022swin} ), and SAM-based models (SAM-Adapter\cite{wu2023medical}, SAMed\cite{zhang2023customized}, AutoSAM\cite{hu2023efficiently}, SAM-LST\cite{chai2024ladder}, D$^2$GPLand\cite{pei2024depth} ). SurgicalSAM, D$^2$GPLand, Deepsolo++, and ABCNet v2 were reimplemented based on their released codes, while the results for other models were directly obtained from \cite{pei2024depth}.

The performance of evaluated models on the L3D test dataset is summarized in Table \ref{tab:comparison_l3d}. Our proposed BCRNet consistently achieves SOTA results across all metrics, outperforming the SOTA method D$^2$GPLand with significant improvements of 5.53\% in DSC, 4.62\% in IoU, and 17.25 pixels in ASSD.
To assess generalization and robustness, we fine-tuned several top-performing models on the P2ILF dataset. As shown in Table~\ref{tab:comparison_p2ilf}, BCRNet again achieved the best performance, surpassing D$^2$GPLand with improvements of 8.23\% in DSC. However, the evaluation results on P2ILF were notably lower than those on L3D, likely due to: (1) the significant domain gap between the datasets, hindering knowledge transfer, and (2) potential inconsistencies in label annotations caused by differences in annotators and landmark definitions.

\begin{figure}[t]
    \centering
    \includegraphics[width=1\textwidth]{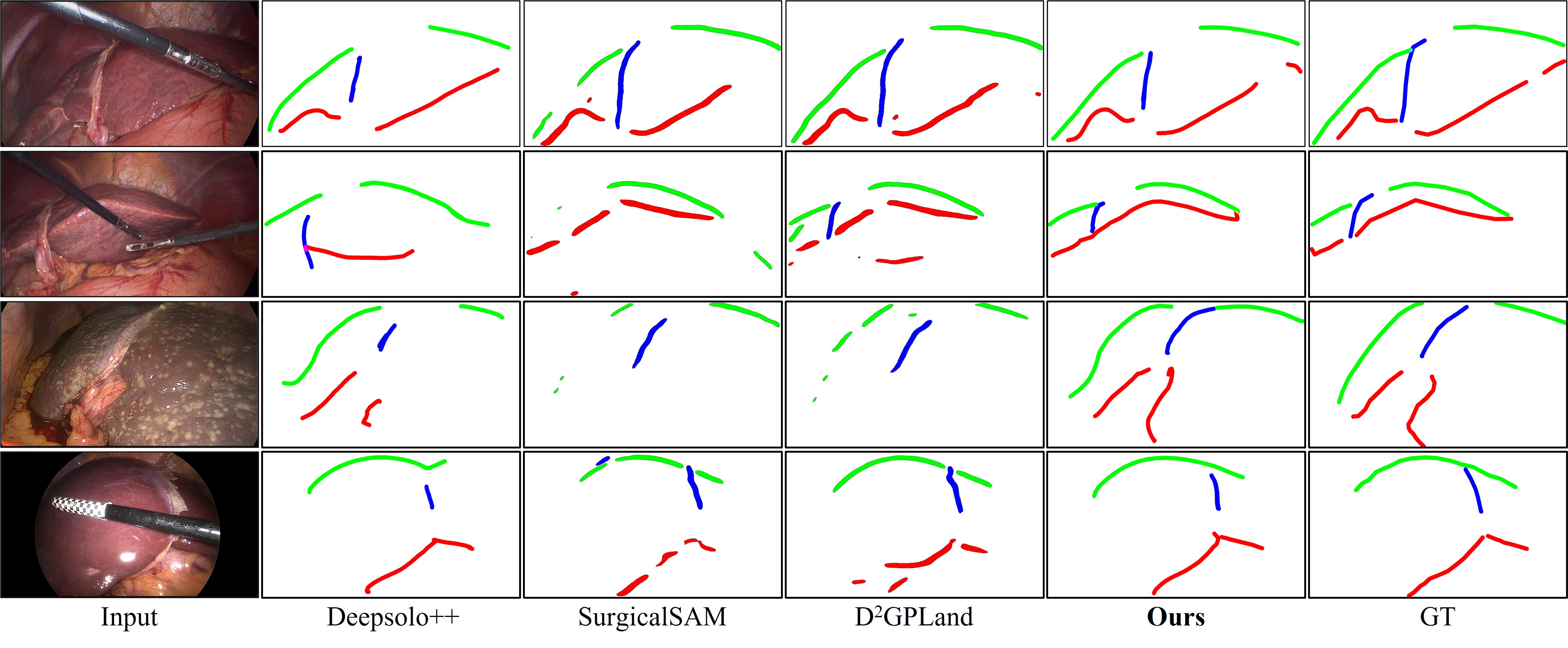}
    \caption{Visualizations of the landmark detection results. The first two rows are from the L3D dataset, and the last two rows are from the P2ILF dataset. }
    \label{fig:visualization}
\end{figure}

Figure~\ref{fig:visualization} presents qualitative comparisons. Segmentation-based methods exhibit issues such as landmark fragmentation, isolated predictions, missing detections, and erroneous outliers. In contrast, our approaches consistently maintain landmark continuity. This robustness not only enhances detection accuracy but also holds significant promise for improving downstream tasks.

\subsection{Ablation Study}

To evaluate the contribution of different components in our proposed method, we conduct an ablation study, with results summarized in Table \ref{tab:ablation}. We design multiple experiments, progressively introducing various modules to observe their impact on the DSC and FWIoU metrics. In the baseline model M.1, no additional modules are utilized, relying solely on CNN layers for feature extraction and simple dense layers for proposal output. This configuration limits the model's ability to capture complex features. Incorporating the MFE module (M.2) leads to improved performance, indicating the positive impact of multi-modal features on model learning. Further introducing the HCR module (M.3) results in a significant enhancement, demonstrating its crucial role in refining feature representation and contextual understanding. Including the ACPI module in M.4 highlights its effectiveness in generating accurate curve proposals, essential for precise predictions. In M.5, removing the MFE module from the full model results in a slight performance decline, underscoring the importance of multi-modal support for the proposed ACPI and HCR modules. Finally, integrating the induction loss allows us to achieve improved performance metrics, showcasing its effectiveness in guiding the model toward better convergence. Overall, the ablation study validates the importance of each component in our approach and their contributions to enhancing overall model performance.

\section{Conclusion}
In this study, we introduce BCRNet, an innovative Bezier curve prediction model designed for  landmark detection in liver laparoscopic images. Our approach pioneers the use of Bezier curves to represent critical anatomical landmarks, leveraging the efficiency and precision of explicit curve modeling to enhance downstream tasks like 2D-3D registration. The framework generates a set of curve proposals for each landmark category and employs a hierarchical optimization process enhanced by deformable attention mechanisms and multi-scale feature integration, ensuring both accuracy and robustness in landmark prediction. Extensive experimental evaluations on the L3D and P2ILF datasets confirm that our method significantly outperforms other SOTA approaches. The exceptional performance of BCRNet underscores its potential as a valuable tool for AR-guided liver surgery and other related medical applications.

\begin{credits}
% \subsubsection{\ackname} A bold run-in heading in small font size at the end of the paper is
% used for general acknowledgments, for example: This study was funded
% by X (grant number Y).

\subsubsection{\discintname}
The authors have no competing interests to declare that are relevant to the content of this article.
\end{credits}

\bibliographystyle{splncs04}
\bibliography{refs}

\begin{thebibliography}{10}
\providecommand{\url}[1]{\texttt{#1}}
\providecommand{\urlprefix}{URL }
\providecommand{\doi}[1]{https://doi.org/#1}

\bibitem{ali2025objective}
Ali, S., Espinel, Y., Jin, Y., Liu, P., G{\"u}ttner, B., Zhang, X., Zhang, L., Dowrick, T., Clarkson, M.J., Xiao, S., et~al.: An objective comparison of methods for augmented reality in laparoscopic liver resection by preoperative-to-intraoperative image fusion from the miccai2022 challenge. Medical image analysis  \textbf{99},  103371 (2025)

\bibitem{blayney2024bezier}
Blayney, H., Tian, H., Scott, H., Goldbeck, N., Stetson, C., Angeloudis, P.: Bezier everywhere all at once: Learning drivable lanes as bezier graphs. In: Proceedings of the IEEE/CVF Conference on Computer Vision and Pattern Recognition. pp. 15365--15374 (2024)

\bibitem{cao2022swin}
Cao, H., Wang, Y., Chen, J., Jiang, D., Zhang, X., Tian, Q., Wang, M.: Swin-unet: Unet-like pure transformer for medical image segmentation. In: European conference on computer vision. pp. 205--218. Springer (2022)

\bibitem{chai2024ladder}
Chai, S., Jain, R.K., Teng, S., Liu, J., Li, Y., Tateyama, T., Chen, Y.w.: Ladder fine-tuning approach for sam integrating complementary network. Procedia Computer Science  \textbf{246},  4951--4958 (2024)

\bibitem{chen2021transunet}
Chen, J., Lu, Y., Yu, Q., Luo, X., Adeli, E., Wang, Y., Lu, L., Yuille, A.L., Zhou, Y.: Transunet: Transformers make strong encoders for medical image segmentation. arXiv preprint arXiv:2102.04306  (2021)

\bibitem{chen2018encoder}
Chen, L.C., Zhu, Y., Papandreou, G., Schroff, F., Adam, H.: Encoder-decoder with atrous separable convolution for semantic image segmentation. In: Proceedings of the European conference on computer vision (ECCV). pp. 801--818 (2018)

\bibitem{feng2022rethinking}
Feng, Z., Guo, S., Tan, X., Xu, K., Wang, M., Ma, L.: Rethinking efficient lane detection via curve modeling. In: Proceedings of the IEEE/CVF Conference on Computer Vision and Pattern Recognition. pp. 17062--17070 (2022)

\bibitem{fuchs1998augmented}
Fuchs, H., Livingston, M.A., Raskar, R., Colucci, D., Keller, K., State, A., Crawford, J.R., Rademacher, P., Drake, S.H., Meyer, A.A.: Augmented reality visualization for laparoscopic surgery. In: Medical Image Computing and Computer-Assisted Intervention—MICCAI’98: First International Conference Cambridge, MA, USA, October 11--13, 1998 Proceedings 1. pp. 934--943. Springer (1998)

\bibitem{hu2023efficiently}
Hu, X., Xu, X., Shi, Y.: How to efficiently adapt large segmentation model (sam) to medical images. arXiv preprint arXiv:2306.13731  (2023)

\bibitem{kirillov2023segment}
Kirillov, A., Mintun, E., Ravi, N., Mao, H., Rolland, C., Gustafson, L., Xiao, T., Whitehead, S., Berg, A.C., Lo, W.Y., et~al.: Segment anything. In: Proceedings of the IEEE/CVF international conference on computer vision. pp. 4015--4026 (2023)

\bibitem{labrunie2023automatic}
Labrunie, M., Pizarro, D., Tilmant, C., Bartoli, A.: Automatic 3d/2d deformable registration in minimally invasive liver resection using a mesh recovery network. In: MIDL. pp. 1104--1123 (2023)

\bibitem{labrunie2022automatic}
Labrunie, M., Ribeiro, M., Mourthadhoi, F., Tilmant, C., Le~Roy, B., Buc, E., Bartoli, A.: Automatic preoperative 3d model registration in laparoscopic liver resection. International Journal of Computer Assisted Radiology and Surgery  \textbf{17}(8),  1429--1436 (2022)

\bibitem{liu2021abcnet}
Liu, Y., Shen, C., Jin, L., He, T., Chen, P., Liu, C., Chen, H.: Abcnet v2: Adaptive bezier-curve network for real-time end-to-end text spotting. IEEE Transactions on Pattern Analysis and Machine Intelligence  \textbf{44}(11),  8048--8064 (2021)

\bibitem{pei2024depth}
Pei, J., Cui, R., Li, Y., Si, W., Qin, J., Heng, P.A.: Depth-driven geometric prompt learning for laparoscopic liver landmark detection. In: International Conference on Medical Image Computing and Computer-Assisted Intervention. pp. 154--164. Springer (2024)

\bibitem{ramalhinho2023value}
Ramalhinho, J., Yoo, S., Dowrick, T., Koo, B., Somasundaram, M., Gurusamy, K., Hawkes, D.J., Davidson, B., Blandford, A., Clarkson, M.J.: The value of augmented reality in surgery—a usability study on laparoscopic liver surgery. Medical Image Analysis  \textbf{90},  102943 (2023)

\bibitem{ronneberger2015u}
Ronneberger, O., Fischer, P., Brox, T.: U-net: Convolutional networks for biomedical image segmentation. In: Medical image computing and computer-assisted intervention--MICCAI 2015: 18th international conference, Munich, Germany, October 5-9, 2015, proceedings, part III 18. pp. 234--241. Springer (2015)

\bibitem{wang2020deep}
Wang, J., Sun, K., Cheng, T., Jiang, B., Deng, C., Zhao, Y., Liu, D., Mu, Y., Tan, M., Wang, X., et~al.: Deep high-resolution representation learning for visual recognition. IEEE transactions on pattern analysis and machine intelligence  \textbf{43}(10),  3349--3364 (2020)

\bibitem{wu2023medical}
Wu, J., Ji, W., Liu, Y., Fu, H., Xu, M., Xu, Y., Jin, Y.: Medical sam adapter: Adapting segment anything model for medical image segmentation. arXiv preprint arXiv:2304.12620  (2023)

\bibitem{xiao2018weighted}
Xiao, X., Lian, S., Luo, Z., Li, S.: Weighted res-unet for high-quality retina vessel segmentation. In: 2018 9th international conference on information technology in medicine and education (ITME). pp. 327--331. IEEE (2018)

\bibitem{ye2023deepsolo}
Ye, M., Zhang, J., Zhao, S., Liu, J., Liu, T., Du, B., Tao, D.: Deepsolo++: Let transformer decoder with explicit points solo for multilingual text spotting. arXiv preprint arXiv:2305.19957  (2023)

\bibitem{yin2022towards}
Yin, W., Zhang, J., Wang, O., Niklaus, S., Chen, S., Liu, Y., Shen, C.: Towards accurate reconstruction of 3d scene shape from a single monocular image. IEEE Transactions on Pattern Analysis and Machine Intelligence (TPAMI)  (2022)

\bibitem{yue2024surgicalsam}
Yue, W., Zhang, J., Hu, K., Xia, Y., Luo, J., Wang, Z.: Surgicalsam: Efficient class promptable surgical instrument segmentation. In: Proceedings of the AAAI Conference on Artificial Intelligence. vol.~38, pp. 6890--6898 (2024)

\bibitem{zhang2023customized}
Zhang, K., Liu, D.: Customized segment anything model for medical image segmentation. arXiv preprint arXiv:2304.13785  (2023)

\bibitem{zhang2022text}
Zhang, X., Su, Y., Tripathi, S., Tu, Z.: Text spotting transformers. In: Proceedings of the IEEE/CVF Conference on Computer Vision and Pattern Recognition. pp. 9519--9528 (2022)

\bibitem{zhou2019unet}
Zhou, Z., Siddiquee, M.M.R., Tajbakhsh, N., Liang, J.: Unet++: Redesigning skip connections to exploit multiscale features in image segmentation. IEEE transactions on medical imaging  \textbf{39}(6),  1856--1867 (2019)

\bibitem{zhudeformable}
Zhu, X., Su, W., Lu, L., Li, B., Wang, X., Dai, J.: Deformable detr: Deformable transformers for end-to-end object detection. arXiv preprint arXiv:2010.04159  (2020)

\end{thebibliography}

\end{document}